\documentclass[12pt, conference]{IEEEtran}
\IEEEoverridecommandlockouts
\usepackage{cite}
\usepackage{amsmath,amssymb,amsfonts}
\usepackage{algorithmic}
\usepackage{graphicx}
\usepackage{textcomp}
\usepackage{url}
\usepackage{booktabs}
\usepackage{hyperref}
\usepackage{tabularx}
\usepackage{comment}
\usepackage{float}
\usepackage[font=footnotesize]{caption}
\usepackage{listings}
\usepackage{xcolor}
\def\BibTeX{{\rm B\kern-.05em{\sc i\kern-.025em b}\kern-.08em
    T\kern-.1667em\lower.7ex\hbox{E}\kern-.125emX}}

\begin{document}
\title{Riazi-8B: An Urdu Large Language Model for Mathematical Reasoning}
%\author{\IEEEauthorblockN{Anonymous Authors}}

%\begin{comment}
\author{

\IEEEauthorblockN{
Azher Ali$^{1}$,
Ibtsam Haider$^{1}$,
Raja Khurram Shahzad$^{2}$,
Seemab Latif$^{1}$,
Mehwish Fatima$^{1,\dagger}$
}

\IEEEauthorblockA{
$^{1}$School of Electrical Engineering and Computer Science (SEECS),\\
National University of Sciences and Technology (NUST), \\
Islamabad, Pakistan\\
\{aali.msds24seecs, iawan.msds24seecs, seemab.latif, mehwish.fatima\}@seecs.edu.pk
}

\IEEEauthorblockA{
$^{2}$Department of Communication, Quality Management and Information Systems,\\
Mid Sweden University, Ostersund, Sweden\\
raja-khurram.shahzad@miun.se
}

\IEEEauthorblockA{
$^{\dagger}$Corresponding author: mehwish.fatima@seecs.edu.pk
}

}
%\end{comment}
\maketitle

% ================================================
\begin{abstract}
Recent LLMs demonstrate strong mathematical reasoning capabilities, but existing gains rely heavily on English-centric training resources and benchmarks. As a result, reasoning performance degrades substantially in low-resource languages such as Urdu, where reasoning-oriented datasets and adapted models remain scarce. Urdu lacks both reasoning-oriented resources and models adapted for multi-step mathematical problem solving, limiting the applicability of recent progress to Urdu-speaking users. We address this gap through Riazi-8B, an Urdu mathematical reasoning model developed through a two-step adaptation process comprising continued pre-training on Urdu Wikipedia and supervised fine-tuning on Urdu Chain-of-Thought data derived from GSM8K. We evaluate Riazi-8B on MGSM-Urdu against existing Urdu instruction-tuned models. Our results show consistent improvements in answer correctness, reasoning quality, response completeness, and Urdu generation. Our findings demonstrate that combining Urdu language adaptation with reasoning-focused fine-tuning is an effective strategy for extending mathematical reasoning capabilities to low-resource languages.
\end{abstract}
 
% ================================== INTRODUCTION SECTION ===================================
\section{Introduction}
\label{sec:intro}
%Problem/ domain definition, need of this problem, connection with Urdu language, and then motivation or application of Urdu Mathematical reasoning. %Challenges associated with Urdu language and reasoning in it.
Mathematical reasoning requires multi-step inference, logical consistency, and the ability to derive verifiable conclusions from intermediate computations~\cite{cobbe2021gsm8k, hendrycks2021math}. Recent advances in Large Language Models (LLMs) substantially improve mathematical reasoning via Chain-of-Thought (CoT) prompting~\cite{wei2022chain} and reasoning-oriented post-training techniques~\cite{shao2024deepseekmath, guo2025deepseekr1}. However, all such efforts are made in high-resourced languages and less to low attention in low-resource languages such as Urdu.

\begin{figure}[t]
    \centering
    \includegraphics[width=\columnwidth]{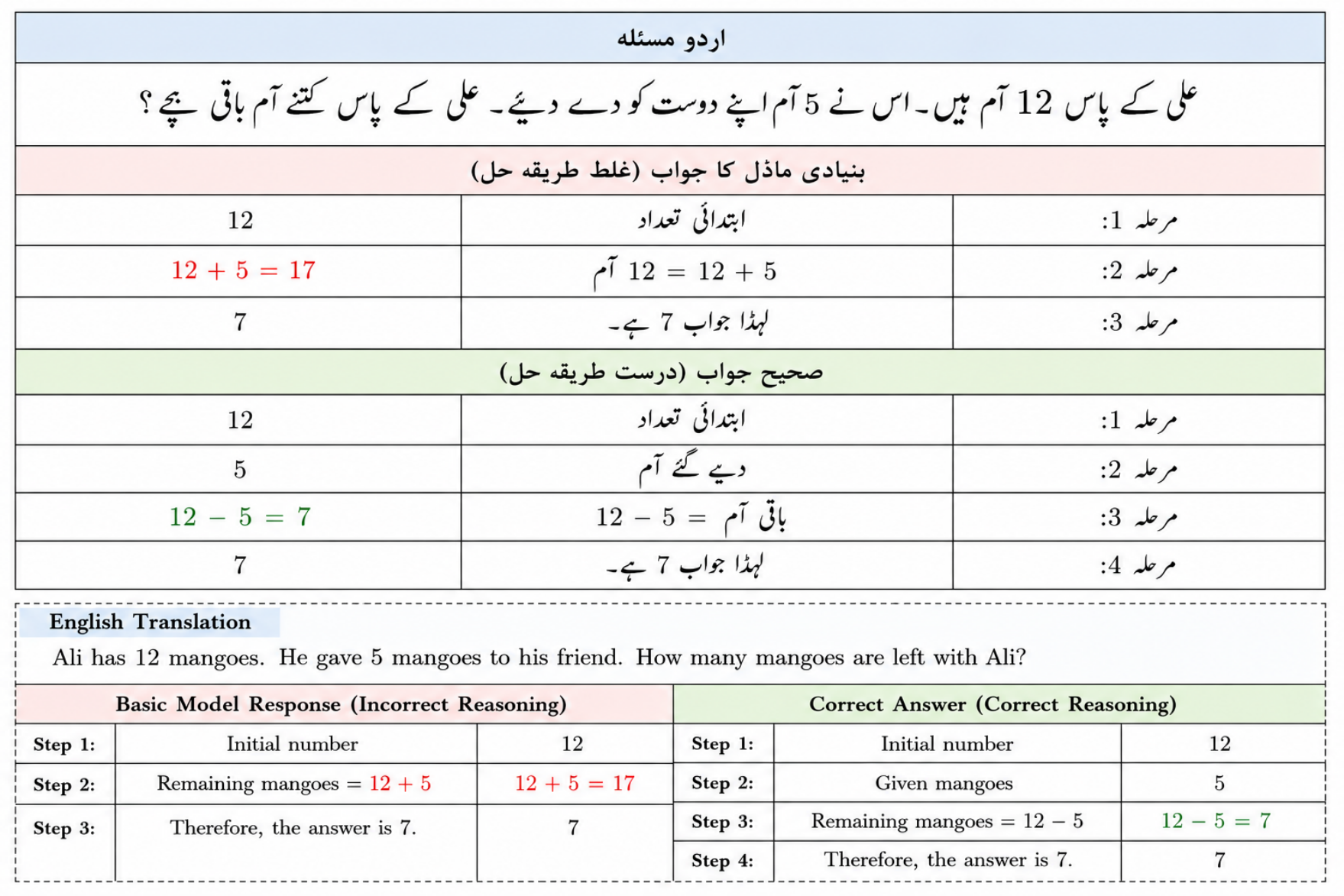}
    \caption{Illustrative Urdu mathematical reasoning example comparing incorrect and correct reasoning traces.}
    \label{fig:motivating_example}
\end{figure}

Urdu presents unique challenges for mathematical reasoning because of its right-to-left Perso-Arabic script, complex word forms, and a lack of reasoning-oriented datasets~\cite{joshi2020state}. Existing Urdu datasets mainly support tasks such as machine translation, sentiment analysis, and question answering, not designed for mathematical reasoning~\cite{jawaid2014urdu, akhter2020sentiment, amjad2020urdu, adeeba2024benchmarking}. While multilingual LLMs can produce fluent Urdu text, they frequently have difficulty with structured multi-step reasoning~\cite{shafique2026urdubench} tasks in Urdu~\cite{ahuja2023mega, bang2023multitask}. Recent models specific to Urdu such as Alif-8B~\cite{shafique2025alif} and Qalb-8B~\cite{hassan2026qalb} often generate fluent responses without maintaining consistent reasoning traces. Figure~\ref{fig:motivating_example} illustrates a representative failure case. The model arrives at the correct answer while producing an incorrect intermediate computation. Such inconsistencies reduce the reliability of generated solutions and limit the usefulness of LLMs in educational and tutoring applications, where reasoning steps are often as important as the final answer.

% propose sol
To the best of our knowledge, there are no prior work to investigate mathematical reasoning in Urdu. We address the gap by investigating mathematical reasoning in urdu. For this, we propose \textit{Riazi-8B}, a two-step Urdu mathematical reasoning framework based on Qwen3-8B\cite{yang2025qwen3}, trained through Continued Pre-Training (CPT) and Supervised Fine-Tuning (SFT) on Urdu mathematical reasoning and Wikipedia \cite{wikimedia2024urdu} datasets to enable step-by-step mathematical reasoning in Urdu.

Our contributions are: (1) We present Riazi-8B, an Urdu large language model designed for mathematical reasoning, (2) We investigate a two-step adaptation strategy that combines Urdu language adaptation with reasoning-focused supervision, and (3) We present an empirical analysis of Urdu mathematical reasoning and provide a comprehensive evaluation framework that combines answer-level metrics with LLM-based assessment of reasoning quality and Urdu fluency.

\begin{figure*}[h]
    \centering
    \includegraphics[width=0.85\textwidth]{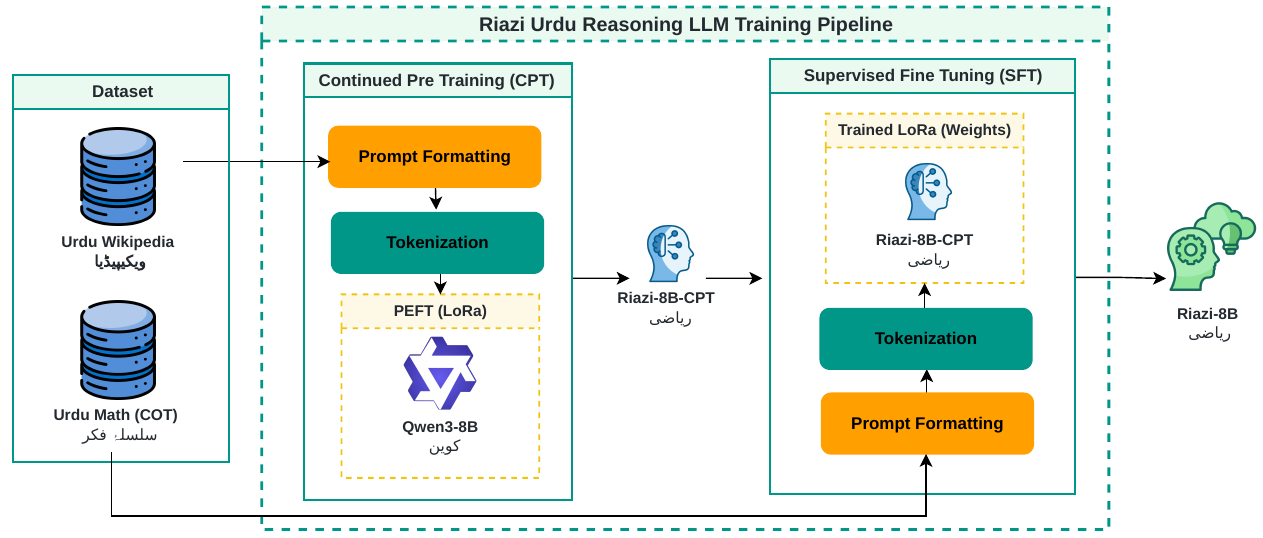}
    \caption{Riazi-8B for Urdu language adaptation through CPT and reasoning-focused SFT.}
    \label{fig:pipeline}
\end{figure*}
% ===================================== RELATED WORK ==================================
\section{Related Work}
\label{sec:related}
We divide this section in three sections, covering mathematical reasoning, Urdu LLMs and low-resource training strategies.

\subsection{Mathematical Reasoning in LLMs}
Datasets such as GSM8K~\cite{cobbe2021gsm8k} and MATH \cite{hendrycks2021math} reveal systematic failures in large models under zero-shot conditions and motivate focus on intermediate reasoning steps. CoT prompting \cite{wei2022chain} shows that supervising full reasoning traces substantially improves performance. Process reward models \cite{lightman2023verify} further demonstrate that step-level supervision outperforms outcome-only supervision. GRPO \cite{shao2024deepseekmath}, a critic-free variant of PPO \cite{schulman2017ppo}, achieves strong mathematical reasoning alignment; DeepSeek-R1 \cite{guo2025deepseekr1} demonstrates that GRPO can incentivize complex multi-step reasoning using only final answer correctness as reward.

\subsection{Urdu LLMs}
Existing Urdu NLP research has focused on tasks like translation \cite{jawaid2014urdu}, sentiment analysis \cite{akhter2020sentiment}, and linguistic processing \cite{amjad2020urdu}  \cite{adeeba2024benchmarking}. Recent Urdu models, including Alif-8B  \cite{shafique2025alif} and Qalb-8B \cite{hassan2026qalb}, use continued pretraining and instruction tuning to improve general Urdu capabilities. However, these models are not specifically designed for mathematical reasoning and rely largely on synthetically generated instruction data, whereas our approach leverages translated human-authored mathematical reasoning. To our knowledge, Riazi-8B is the first Urdu-focused LLM trained explicitly for step-by-step mathematical reasoning.

\subsection{Low-Resource Training Strategies} 
Joshi et al.\ \cite{joshi2020state} classify languages by digital resource volume, showing that most fall in categories with minimal NLP representation. MEGA \cite{ahuja2023mega} and Bang et al.\ \cite{bang2023multitask} confirm that reasoning performance degrades substantially in low-resource settings, directly motivating language-specific training. Domain-adaptive pre-training \cite{gururangan2020dont} provides the empirical basis for CPT: continued training on domain-specific text yields consistent gains, particularly when the target domain is distant from the pretraining distribution, precisely the case for Urdu. 

% ================================================
\section{Proposed Solution: Riazi-8B}\label{sec:method}
We propose Riazi-8B, an Urdu mathematical reasoning pipeline built on Qwen3-8B~\cite{yang2025qwen3}. The pipeline combines CPT on Urdu Wikipedia with SFT on Urdu Chain-of-Thought mathematical reasoning data. We use Qwen3-8B as our base model due to its multilingual capabilities. Both adaptation steps use LoRA~\cite{hu2022lora} for parameter-efficient training. Figure~\ref{fig:pipeline} summarizes the overall pipeline\footnote{Code and training resources will be released upon acceptance.}. %\footnote{\href{https://github.com/azheraly/Riazi-8B-First-Urdu-Reasoning-Large-Language-Model}{We release our code publicly under the license CC TODO: ADD the license and proper statement}}.

\subsection{Continued Pre-Training (CPT)} \label{subsec:cpt} 
Our first step is CPT of Qwen3-8B on Urdu Wikipedia. Our objective is to adapt Urdu fluent generation before introducing mathematical reasoning supervision. We use LoRA~\cite{hu2022lora} for parameter-efficient adaptation and update only a small subset of trainable parameters.
 
\subsection{Supervised Fine-Tuning}\label{subsec:sft}
Following CPT, our second step fine-tunes the adapted model on GSM8K-Urdu. Each training instance consists of an Urdu mathematical problem and a corresponding reasoning trace with the final answer. With this step, we introduce Urdu mathematical reasoning in the trained model. During inference, we use temperature 0.6, top-$p$ 0.95, top-$k$ 20, and a maximum sequence length of 2,048 tokens. Table~\ref{tab:experimental-setup} summarizes experimental setup for both stages. We use LoRA~\cite{hu2022lora} for both CPT and SFT to enable parameter-efficient adaptation without updating the full model parameters.

\begin{table}[ht]
\caption{Model and configuration details for the CPT and SFT stages, including decoding settings used during evaluation.}
\label{tab:experimental-setup}
\centering
\footnotesize    % or \scriptsize for even smaller font
\renewcommand{\arraystretch}{1.0}
\begin{tabular}{lcc}
\toprule
\textbf{Hyperparameter} & \textbf{CPT} & \textbf{SFT} \\
\midrule
Base model          & Qwen3-8B & Riazi-8B-CPT \\
Learning rate       & $5\times10^{-5}$ & $5\times10^{-5}$ \\
Embedding LR        & $1\times10^{-5}$ & $1\times10^{-5}$ \\
LR schedule         & Cosine & Cosine \\
Epochs              & 1 & 2 \\
%Batch size          & 64 & 64 \\
%LoRA rank ($r$)     & 128 & 128 \\
%LoRA $\alpha$       & 32 & 32 \\
%Precision           & bfloat16 & bfloat16 \\
%Max sequence length & 2048 & 2048 \\
%Temperature (inf.)  & 0.6 & 0.6 \\
%Top-$p$ (inf.)      & 0.95 & 0.95 \\
%Top-$k$ (inf.)      & 20 & 20 \\
%Random seed         & 3407 & 3407 \\
\bottomrule
\end{tabular}
\end{table}

%====================================================== Experiments =====================================
\section{Experiments}\label{sec:experiments}

\subsection{Datasets}
We use Urdu Wikipedia~\cite{wikimedia2024urdu} for continued pretraining, GSM8K-Urdu~\cite{shafique2026urdubench} for supervised fine-tuning, and MGSM-Urdu~\cite{shafique2026urdubench} for evaluation. Urdu Wikipedia contains approximately 200,000 articles and serves as the linguistic adaptation corpus. GSM8K-Urdu is an Urdu translation of GSM8K~\cite{cobbe2021gsm8k} containing approximately 8,500 arithmetic word problems paired with step-by-step solutions and final answers. MGSM-Urdu contains 250 held-out grade-school mathematics problems and is used to evaluate generalization beyond the training data. We perform language verification using fastText~\cite{joulin2016fasttext} and apply Unicode normalization, document-level de-duplication, quality filtering, and tokenizer compatibility checks using the Qwen3-8B tokenizer~\cite{yang2025qwen3}.

\subsection{Models}
We compare Riazi-8B with three baseline models: (1) Qalb-8B~\cite{hassan2026qalb}, (2) Alif-8B~\cite{shafique2025alif}, and (3) Llama-8B~\cite{touvron2023llama}. Qalb-8B and Alif-8B represent existing Urdu-adapted LLMs, while Llama-3.1-8B-Instruct provides a strong multilingual baseline at a comparable scale. We evaluate all models using identical decoding parameters (temperature $=0.6$, top-$p=0.95$, top-$k=20$, maximum sequence length $=2048$ tokens).

%========================================================
\begin{figure}[h]
    \centering
    \includegraphics[width=0.6\columnwidth]{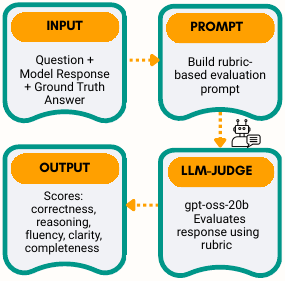}
    \caption{Evaluation pipeline where model responses are assessed against ground truth using an LLM judge.}
    \label{fig:llm_judge}
\end{figure}
% ================================================
\subsection{Evaluation}
We report three metrics: Exact Match Accuracy (EM) measures the fraction of problems for which the extracted numerical answer matches the reference exactly; this is our primary metric. Urdu Output Purity (UOP) measures the fraction of generated tokens belonging to Urdu script, identified via fastText \cite{joulin2016fasttext}, capturing the language collapse failure mode. Step Completeness Score (SCS) measures whether the model produces at least two distinct intermediate reasoning steps before the final answer, distinguishing structured CoT generation from direct retrieval. These three metrics characterize answer correctness, language fidelity, and reasoning structure, exposing each failure mode independently.

\subsection{LLM-as-Judge}\label{subsec:llm-judge}
To evaluate the quality of generated responses, we apply a rubric-based evaluation framework grounded in the LLM-as-Judge paradigm~\cite{zheng2023judging, liu2023calibrating}. As illustrated in Figure~\ref{fig:llm_judge}, we use \texttt{gpt-oss-20b} as an automated evaluator. The judge receives the Urdu problem, the model's complete response, and the reference answer, and scores the response on five dimensions using a 5-point Likert scale: (i) \textit{Correctness}: accuracy of the final answer and intermediate conclusions, (ii) \textit{Reasoning}: coherence and validity of the reasoning process, (iii) \textit{Urdu Fluency}: grammatical correctness and naturalness of the generated Urdu text, (iv) \textit{Clarity}: absence of ambiguity in the reasoning and conclusions, and (v) \textit{Completeness}: coverage of all necessary reasoning steps. We report the mean and standard deviation for each dimension across the evaluation set. These metrics complement exact-match accuracy by capturing qualitative aspects of reasoning and response quality.

%========================================================

\begin{table}[t]
\caption{Evaluation results on Urdu MGSM: EM, UOP, and SCS.}
\label{tab:main-results}
\centering
\footnotesize
\renewcommand{\arraystretch}{1.05}

\begin{tabular}{lcccc}
\toprule
\textbf{Metrics} & \textbf{Alif-8B} & \textbf{Qalb-8B} & \textbf{Llama} & \textbf{Riazi-8B} \\
\midrule
EM $\uparrow$ & 29.2 & 32.0 & 59.6  & \textbf{64.4} \\
UOP $\uparrow$ & 70.4 & 63.0 & 77.0  & \textbf{82.7} \\
SCS $\uparrow$  & 53.4 & 53.4 & 77.2 & \textbf{83.8} \\
\bottomrule
\end{tabular}
\end{table}

% ================================================ Results and Discussion ==============================
\begin{table*}[h]
\caption{LLM-as-a-Judge evaluation results on the Urdu reasoning benchmark. Scores are reported as mean $\pm$ standard deviation. }
\label{tab:llm-judge}
\centering
\footnotesize
\renewcommand{\arraystretch}{1.05}

\begin{tabular}{lcccc}
\toprule
\textbf{Metrics} & \textbf{Alif-8B} & \textbf{Qalb-8B} & \textbf{Llama-8B} & \textbf{Riazi-8B} \\
\midrule
Correctness $\uparrow$ &
2.27 $\pm$ 1.84 &
2.27 $\pm$ 1.84 &
3.51 $\pm$ 1.92 &
\textbf{3.87 $\pm$ 1.77} \\

Reasoning $\uparrow$ &
2.23 $\pm$ 1.77 &
2.29 $\pm$ 1.78 &
3.50 $\pm$ 1.85 &
\textbf{3.90 $\pm$ 1.69} \\

Urdu Fluency $\uparrow$ &
3.52 $\pm$ 1.02 &
3.15 $\pm$ 1.01 &
3.85 $\pm$ 0.89 &
\textbf{4.14 $\pm$ 0.80} \\

Clarity $\uparrow$ &
2.83 $\pm$ 1.52 &
2.72 $\pm$ 1.61 &
3.84 $\pm$ 1.51 &
\textbf{4.20 $\pm$ 1.29} \\

Completeness $\uparrow$ &
2.67 $\pm$ 1.62 &
2.67 $\pm$ 1.70 &
3.86 $\pm$ 1.54 &
\textbf{4.19 $\pm$ 1.34} \\

\midrule
\textbf{Mean} &
2.70 &
2.62 &
3.71 &
\textbf{4.06} \\
\bottomrule
\end{tabular}
\end{table*}

\section{Results}\label{sec:results}
Table~\ref{tab:main-results} reports the evaluation results on MGSM-Urdu. Riazi-8B achieves the highest performance across all three metrics, outperforming both Urdu-specific and multilingual baselines. The strongest baseline, Llama, already exhibits competitive mathematical reasoning performance, achieving 59.6\% exact-match accuracy and 77.2\% step completeness. However, Riazi-8B improves upon these results, reaching the highest answer accuracy, Urdu output purity, and reasoning completeness.

The gap is most pronounced for Urdu generation quality. While Llama achieves relatively strong reasoning performance, its Urdu output purity remains below that of Riazi-8B. In contrast, Riazi-8B combines competitive reasoning with more consistent Urdu generation, suggesting that language adaptation contributes beyond answer correctness alone. The improvements in SCS further indicate that the model produces more complete reasoning traces before arriving at the final answer. Together, these results suggest that combining Urdu language adaptation with reasoning-focused supervision improves both reasoning quality and language fidelity.

%========================================================

\subsection{Response Quality Analysis}
Table~\ref{tab:llm-judge} reports the LLM-as-Judge evaluation across five quality dimensions. Riazi-8B achieves the highest scores on Correctness, Reasoning, Urdu Fluency, Clarity, and Completeness, indicating improvements beyond answer-level accuracy. The strongest baseline, Llama-8B, performs competitively across all dimensions, yet Riazi-8B maintains a consistent advantage. The largest gains appear in Completeness and Urdu Fluency, suggesting that the model produces more complete reasoning traces while maintaining higher-quality Urdu generation.

%========================================================
These results indicate that language adaptation alone is insufficient for mathematical reasoning. Although Llama-8B demonstrates strong general reasoning capabilities, reasoning-focused supervision further improves correctness, clarity, and completeness in Urdu. In contrast, Alif-8B and Qalb-8B lag substantially behind on both reasoning-related and language-related dimensions, highlighting the difficulty of transferring mathematical reasoning to Urdu without task-specific adaptation.

\section{Limitations} 
The continued pretraining data mainly comes from Urdu Wikipedia, which has little content in mathematics and science. This limits the model's ability to gain domain-specific reasoning skills. Second, the instruction-tuning data comes from GSM8K and focuses mostly on grade-school math. This leaves the model's performance in advanced mathematics unmatched. Third, because the Urdu instruction-tuning dataset is relatively small, the model may forget some previously learned English skills. This would lower its performance in multiple languages. Finally, evaluation is only done on the 250-problem MGSM-Urdu benchmark. Having larger and more varied Urdu reasoning benchmarks would allow for a better assessment of the model's abilities.

% ===================================== Conclusion =======================================

\section{Conclusions}\label{sec:conclusion}
We present Riazi-8B, the first LLM specifically trained to produce complete step-by-step mathematical reasoning in Urdu. It is a two-step Urdu mathematical reasoning system built using CPT, followed by CoT instruction tuning. The proposed approach improves both linguistic alignment and structured reasoning in Urdu using a parameter-efficient LoRA-based training pipeline. Experimental results on MGSM-Urdu show that Riazi-8B consistently outperforms strong Urdu instruction-tuned baselines, achieving higher exact match accuracy, Urdu output purity, and step completeness. These results confirm that combining language adaptation with reasoning-focused fine-tuning is effective for low-resource mathematical reasoning tasks.

\bibliographystyle{IEEEtran}
\bibliography{references}

\end{document}